# The Bounded Bayesian

**Kathryn Blackmond Laskey**
Department of Systems Engineering and $C^3I$ Center
George Mason University
Fairfax, VA 22030
klaskey@gmu.edu

## Abstract

The ideal Bayesian agent reasons from a global probability model, but real agents are restricted to simplified models which they know to be adequate only in restricted circumstances. Very little formal theory has been developed to help fallibly rational agents manage the process of constructing and revising small world models. The goal of this paper is to present a theoretical framework for analyzing model management approaches. For a probability forecasting problem, a search process over small world models is analyzed as an approximation to a larger-world model which the agent cannot explicitly enumerate or compute. Conditions are given under which the sequence of small-world models converges to the larger-world probabilities.

## 1 INTRODUCTION

This paper is concerned with the use of decision theory as a framework for analyzing the behavior of an intelligent autonomous agent. It is generally agreed that our best exemplars of intelligent autonomous agency--human beings--are not Bayesians. It is also generally agreed that many systematic departures from the Bayesian norm can be explained as heuristics which people use to cope with computational constraints. Given that any agent we can build will have limited computational capacity, what role should decision theory play for the enterprise of developing intelligent autonomous agents?

Bayesian decision theory justifies many qualitative features of generally accepted "good" reasoning, and explains many qualitative features of "bad" reasoning (see, e.g., Pearl, 1988; Howsen and Urbach, 1989). The theory is often applied in cases in which the idealized axioms are clearly unrealistic. But successful decision analysis usually requires a highly trained (human) decision analyst working with the decision maker to construct a model of the problem. Similarly, decision theoretic computer systems either are hand-crafted for constrained, well-understood domains, or use heuristic methods at the meta level to construct an object-level decision theoretic model (e.g., Andreassen, et al., 1987; Goldman and Charniak, 1990).

It has been argued that the latter is the appropriate role for decision theory (Wellman and Heckerman, 1987). A heuristic meta-level reasoner constructs decision theoretic models for "small worlds," or restricted sets of propositions and actions. Decision theory enforces consistency on beliefs and action recommendations produced by the "small-world" models. The meta-level reasoner is free to use the output of a small world model, modify it, or change the model, depending on the situation. But this approach begs two important questions. First, if formal justifications of decision theory assume "grand world" coherence, what justifies decision theory on the small world? Second, what theory underlies the meta-reasoner?

This paper is part of a research effort to develop a decision theoretically based foundation for the meta-reasoner. My basic formulation can be summarized as follows. The meta-reasoner controls a search process over a space of small world models. This search process is viewed as an approximation to a (computationally infeasible) decision theoretic model on some larger world. I assume that the agent would prefer the larger world model, if it could be computed, to the feasible small world approximation. Within this formal framework, search processes over small worlds can be analyzed to determine how well they perform as approximations to the larger world model. Note that I do *not* assert that the larger world model represents the agent's "true" grand-world beliefs--indeed, I do not even assume that the agent has grand-world beliefs.

In this paper, I restrict attention to the common and important class of problems in which an agent forecasts probabilities of a sequence of events. These events may be repeated events of the same kind, as when an agent assesses probabilities for different diseases given patients' observed symptoms. Alternatively, the events may be of different kinds, as when a common sense reasoner makes predictions about events it expects in its current situation. In either case, I assume the agent receives feedback about whether the forecast events actually occur. The agent can search a countably infinite space of probability models, but has access to only two models (the current model and a comparison model) at any given time. Under given conditions on the agent's search process, I derive results about convergence of the agent's model to the larger world probabilities.

## 2 THE SEQUENTIAL FORECASTING PROBLEM

Consider an agent attempting to forecast the probabilities of a sequence $E_1, E_2, \ldots$ of arbitrary events occurring sequentially in time.[1] The events can be defined in advance or defined dynamically on the basis of information observed at times prior to time t.

The agent's task is to state a probability $p_t$ for $E_t$. This forecast may take into account the agent's entire experience up to, but not including, the events which occur at time period t. Formally, let $\mathcal{H}_t$ be a $\sigma$-field representing all events known to the agent at times up to and including time period t. The agent's knowledge increases monotonically with time--that is, $\mathcal{H}_0 \subset \mathcal{H}_1 \subset \ldots$ The condition that the forecast must make use only of information known before time t is formally expressed by requiring that the agent's forecast $p_t$ be an $\mathcal{H}_{t-1}$ measurable function. Thus,

$$p_t = P(E_t \mid \mathcal{H}_{t-1}), \quad \text{where } \mathcal{H}_0 \subset \mathcal{H}_1 \subset \ldots \quad (1)$$

expressing dependence of the agent's probability estimate for $E_t$ on the past history $\mathcal{H}_{t-1}$ of observations.

Denote the indicator function of $E_t$ by $X_t$:

$$X_t = \begin{cases} 1 & \text{if } E_t \\ 0 & \text{if } \bar{E}_t \end{cases} \quad (2)$$

The expression $P(X_t \mid \mathcal{H}_{t-1})$ is used to denote $P(E_t \mid \mathcal{H}_{t-1})$ when $X_t = 1$ and $P(\bar{E}_t \mid \mathcal{H}_{t-1})$ when $X_t = 0$.

## 3 MODEL SEARCH AND REVISION

### 3.1 ASYMPTOTIC BEHAVIOR OF THE MIXTURE MODEL

Consider a set of models indexed by a parameter $\omega \in \Omega$. Each model represents a recipe for computing forecasts $P_\omega(E_t \mid \mathcal{H}_{t-1})$ at each time period from information known to the agent prior to that time. The set $\Omega$ can represent either a space of implicitly defined models which a single agent can enumerate, or a population of agents, each of whom has a recipe for predicting $E_t$. This paper considers the probability forecasts of a single agent who has access to the models in $\Omega$, either because she is the agent enumerating and computing the implicitly defined models, or because she can ask the other agents in the population for their forecasts. However, the agent is assumed to have finite resources for computing or communicating, and cannot access all the models at once.

For this paper, I assume that the parameter space $\Omega$ is finite. The issues involved in extending to the countable case are discussed briefly. This assumption is not so restrictive as it may seem, if the $\omega$ are thought of as indexing different *model structures*, whose parameters can be estimated given data. The *parameter space* for any given $P_\omega$ may be uncountable. For example, $\omega$ might index a model in which the probability of $E_t$ depends on a set of "symptoms" $(Y_{1t}, \ldots Y_{rt})$ observable prior to time t, where the unknown probabilities $\pi(Y_{1t}, \ldots, Y_{rt})$ are estimated from the data, and could have any values between 0 and 1. There is no requirement that the models be explicitly Bayesian; $P_\omega$ can be any recipe for computing a degree of belief in $E_t$ given information available prior to time t.

The set $\Omega$ comprises all the models the agent is able to consider. For the present, let us assume that the agent believes $\Omega$ to be adequate. That is, the agent's larger-world model is a probability mixture over the $P_\omega$. An agent would assign a probability mixture over the $P_\omega$ if she believed one of the models to be correct, and expressed her uncertainty about which one was correct as a probability distribution over $\Omega$. Note, however, that an agent might be willing to assign a mixture distribution over the $\Omega$ without believing in the literal truth of any of the $P_\omega$. Note also that the agent may not be able to compute this mixture distribution--but she believes that if she could compute it, the mixture would provide accurate probability forecasts.

Let the agent's prior probability for the model $\omega$ be denoted by $\alpha_{\omega 0} > 0$, where $\Sigma \alpha_{\omega 0} = 1$. Define

$$\alpha_{\omega t} = \frac{P_\omega(X_t \mid \mathcal{H}_{t-1})\alpha_{\omega, t-1}}{\sum_\omega P_{\omega'}(X_t \mid \mathcal{H}_{t-1})\alpha_{\omega', t-1}}. \quad (3)$$

The quantity (3) is the agent's posterior probability for model $\omega$ given the data $X_1, X_2, \ldots, X_t$. The agent's probability for event $E_t$ given all information available prior to time t is then given by:

$$P(E_t \mid \mathcal{H}_{t-1}) = \sum_\omega \alpha_{\omega, t-1} P_\omega(E_t \mid \mathcal{H}_{t-1}). \quad (4)$$

Adopt for the moment the position of an agent who believes that one of the models is correct, although she does not know which one. Call this unknown correct model $P_{\omega^*}$. The following lemma shows that the agent believes that the mixture distribution will eventually arrive at probabilities which agree with those of the unknown correct model $P_{\omega^*}$.

*Lemma 1*: With probability 1 under the distribution $P_{\omega^*}$, the posterior probability $\alpha_{\omega t}$ of model $P_\omega$ converges to a limiting value $\alpha_\omega$ as $t \to \infty$. If the limiting posterior probability is greater than zero, then

---

[1] The exposition is limited to binary events, but the results remain valid for events with an arbitrary finite number of outcomes (as when forecasting probabilities of different diseases).

$$\frac{P_\omega(X_t \mid \mathcal{H}_{t-1})}{P_{\omega*}(X_t \mid \mathcal{H}_{t-1})} \to 1 \quad (5)$$

*Corollary 2*: With probability 1 under $P_{\omega*}$, $P(X_t \mid \mathcal{H}_{t-1}) - P_{\omega*}(X_t \mid \mathcal{H}_{t-1}) \to 0$ as $t \to \infty$.

Corollary 2 states that if the data are produced by some unknown "true" probability distribution, then Bayesian updating of a mixture distribution which assigns nonzero probability to the correct model will eventually result in forecasts agreeing with the true probabilities.[2]

### 3.2 SEARCH AND REVISION - NO FORGETTING

Suppose that (4) describes the agent's beliefs about the $X_t$. In other words, the agent's beliefs can be represented as a mixture over the models $P_\omega$. However, the agent cannot compute the full mixture distribution, and thus the agent does not "know" her larger-world beliefs at any point in time.

In this section, I analyze the behavior of a heuristic model search and revision process, referred to as SR. Informally, the SR process can be described as follows. At each time period, the agent makes a prediction given her current model (one of the $P_\omega$), and at the same time (i.e., based on the same information) selects an alternate model according to a model search distribution. She then compares the relative posterior probabilities of the current and alternate models. If the posterior probability of her current model is higher than that of the alternate model, she retains the current model; otherwise, she replaces it with the alternate model.

Formally, let $\omega_{ct}$ and $\omega_{at}$ denote the current and alternate models at time period t. Let $\mu_t(\omega)$ denote the probability at time t of selecting model $\omega$ as the new alternate model. The initial models $\omega_{c0}$ and $\omega_{a0}$ may be chosen arbitrarily. The model search distribution $\mu_t(\omega)$ is a real-valued $\mathcal{H}_{t-1}$-measurable function such that $\mu_t(\omega) > 0$ for all $\omega \in \Omega$ and $\Sigma \mu_t(\omega) = 1$. (The neasurability condition means that the probability of selecting model $\omega$ depends only on $\mathcal{H}_{t-1}$, that is, on information known to the agent prior to time t.) For each $t > 0$, $\omega_{at}$ is assumed to be selected independently of $X_t$ according to the probability distribution $P[\omega_{at}=\omega \mid \mathcal{H}_{t-1}] = \mu_t(\omega)$. Now, suppose that the agent updates her model by selecting for the next period's current model the model with the higher posterior probability given all information observed so far. That is, suppose that $\lambda_{\omega t}$ denotes the likelihood of the observed data under the model indexed by $\omega$ (an explicit formula is given as Equation (A-1) in the appendix below). Then the agent selects her current model at time t according to:

$$\omega_{ct} = \begin{cases} \omega_{c,t-1} & \text{if } \alpha_{\omega_{c,t-1}} 0 \lambda_{\omega_{c,t-1} t} \\ & \geq \alpha_{\omega_{a,t-1}} 0 \lambda_{\omega_{a,t-1} t} \quad (6) \\ \omega_{a,t-1} & \text{otherwise .} \end{cases}$$

It is assumed that $\omega_{ct}, \omega_{at} \in \mathcal{H}_t$. That is, the agent's knowledge at the time of forecasting includes knowledge of which model is being used for forecasting and which is the alternate model. This assumption allows the model search probabilities $\mu_t$ to depend on the previous search path as well as on other information known to the agent prior to time t. Apart from $\mathcal{H}_{t-1}$-measurability and the condition of Theorem 3 below, no restrictions are placed on the model search distribution $\mu_t$. Analyzing the behavior (especially short-term behavior) of SR under different types of model search distribution is an interesting research problem, but beyond the scope of the present paper.

The agent's forecast at time t is given by

$$p_t = P_{\omega_{ct}}(X_t \mid \mathcal{H}_{t-1}). \quad (7)$$

Note that (6) uses relative posterior probabilities, so that the agent need not compute the normalizing factor in the denominator of (3) (computing this would require her to know $\lambda_{\omega t}$ for all $\omega$, which I have assumed is infeasible). Nevertheless, the agent does need to compute $\lambda_{\omega_{a} t}$, the likelihood under the new model of the entire past sequence of data (the reasonableness of this assumption is discussed in greater detail below).

The SR process makes no claim to optimality, nor is it intended as a faithful psychological account of how people reason. It is put forward as a simple procedure which plausibly accounts for important qualitative features of human scientific reasoning, such as belief persistence (adherence to the current hypothesis despite disconfirming evidence) and sudden "paradigm shifts" (which will occur when the search process visits a model more probable than the current one).

Suppose, for example, that if $\omega$ is near $\omega'$ (by some appropriate proximity measure), the probability distributions $P_\omega$ and $P_{\omega'}$ are similar. Suppose also that the model search distribution places high probability on $\omega_{at}$ near the current model $\omega_{ct}$. (This corresponds to the heuristic judgment that if a model fits well, similar models can be expected to fit better than alternate models chosen at random.) Most modifications to the agent's current theory, then, are likely to be relatively minor. If the current theory is not correct, however, occasional "anomalies" can be expected. That is, events an incorrect theory considers improbable might be expected to occur more often than the theory predicts. Eventually, a distant model may be generated that has higher posterior probability than the current model. This will be the case if the new model explains anomalies in the current model, approximates the current model on the things the current model predicts well, and does not have too low a prior probability relative to the current model. (Note that *ad hoc* hypotheses constructed specifically to fit the anomalies will fail to be accepted because they have low prior probabilities.) When a better model is found, SR replaces $\omega_{ct}$ with the new model $\omega_{at}$, and a paradigm shift occurs. The search process now begins searching for models near the new model. Because it is unlikely that the best model in that new neighborhood was enumerated initially, there ensues a period in which the new model is incrementally refined by searching its neighborhood for improved models. Thus, SR with an appropriately defined model search process can produce behavior which mimics qualitatively the process of scientific inference put forward by Thomas Kuhn (1962).[3]

---
[2] Proofs are given in the Appendix. The proofs of Lemma 1 and Corollary 2 carry through for countable $\Omega$ as well.

The following results compare the asymptotic behavior of SR with that of the mixture distribution. For Theorem 3, again assume that the agent believes one of the models, denoted by $P_{\omega^*}$ is correct.

*Theorem 3:*   Suppose that $\omega_{at}$ visits each $\omega \in \Omega$ infinitely often. Then under $P_{\omega^*}$,

$$\frac{P_{\omega_{ct}}(X_t \mid \mathcal{H}_{t-1})}{P_{\omega^*}(X_t \mid \mathcal{H}_{t-1})} \to 1. \qquad (8)$$

with probability 1 as $t \to \infty$.

Corollary 4 drops the reference to an unknown correct distribution.

*Corollary 4:*   Under the mixture distribution, the probability that

$$\mid P_{\omega_{ct}}(X_t \mid \mathcal{H}_{t-1}) - P(X_t \mid \mathcal{H}_{t-1}) \mid > \varepsilon \qquad (9)$$

converges to zero for each $\varepsilon > 0$.

Corollary 4 states that under the mixture distribution, the difference between the forecast of $\omega_{ct}$ and the forecast of the mixture distribution *converges in probability* to zero. This means that the probability that this difference exceeds in absolute value any fixed nonzero constant becomes arbitrarily small as t approaches infinity. In other words, with more observations it becomes increasingly probable that the current model's forecast will be very close to the mixture model's forecast.

The SR procedure is intended to model an agent restricted in her ability to represent and compute her entire mixture distribution. But it is deficient in one important respect as a plausible model for a bounded agent. Each time the agent enumerates a new model she must compute the likelihood of the entire sequence of previously observed data under the new model. This means that the agent must not only remember the entire past sequence of data points $X_t$, but must also remember all past values of any auxiliary variables $Y_{1t}, Y_{2t}, \ldots$ that might be required to compute any of the $P_\omega$. Finally, to compute $\lambda_{\omega_{at}}$ she must compute and multiply together the values $P(X_s \mid \mathcal{H}_{s-1})$ for all $s < t$. The feasibility of this is quite doubtful. It may be feasible for the agent to use a heuristic approximation of $\lambda_{\omega_{at}}$. Of course, the results of this section apply only when the correct likelihoods are used.

---

[3]The term paradigm shift is usually reserved for major belief shifts, typified by the replacement of Newton's theory by the theory of relativity. The formal framework presented here applies whether $\omega$ indexes grand theories of the universe or simple time series models of stock market prices. Although one might balk at using the term paradigm shift in the latter context, it is interesting to note that the same formal framework applies to belief changes at a variety of levels.

The SR procedure is feasible as a model of information exchange among rational agents. For suppose that the set $\Omega$ indexes a large population of different agents. Each agent $\omega$ has his own model $P_\omega$. After each trial, each agent computes his own model's likelihood $\lambda_{\omega t}$ for the data. (This is simply the product of $\lambda_{\omega,t-1}$ and the probability forecast for the event which actually occurred at period t.) The SR agent begins by selecting one of the agents $\omega_{ct}$ as an advisor. On each trial, she interviews an alternate advisor $\omega_{at}$, and replaces her current advisor if she believes it is more probable that the interviewee will provide correct predictions than that the current advisor will. Theorem 1 says that such a sequential selection process will yield asymptotically correct forecasts if one of the agent's models is asymptotically correct and each agent is interviewed infinitely often.

SR possesses the important property that it will eventually track correctly any process which one of the models in the search space can track. This is true assuming only that each model is visited infinitely often. In particular, the probability that the correct model $\omega^*$ is visited on any trial can be very small, and can even decrease with t (as long as it does not decrease so rapidly that $\omega^*$ is not visited infinitely often).

But real forecasters cannot wait until infinity. Of more importance to them is the shorter-term performance of SR. This will depend on the properties of the model enumeration process $\mu_t$, of which I have said nothing but that it visits each model infinitely often. For good short-term forecasts, the forecaster requires good diagnostics for suggesting new models to enumerate based on performance of the current model (cf., Laskey, 1991).

### 3.3 RELAXING THE MEMORY REQUIREMENT

Usually when entertaining a new theory one does not have available all the data which went into the computation of the current theory's likelihood function. If one moves from the population of forecasters interpretation to one in which a single forecaster enumerates models sequentially, the memory requirement of Section 3.2 becomes untenable. This section considers relaxation of the memory assumption.

To do this, I define a search strategy which can be followed by a memory limited agent. This strategy will be called SRF (for search and revise with forgetting). SRF works like SR in that the agent makes predictions on the basis of a current model while searching a space of alternate models. Again, the current model and the alternate model are compared on the basis of relative posterior probabilities, and the more probable model is retained. The differences between SRF and SR are:

- Posterior probabilities are computed on the basis of data observed after the alternate model has been enumerated; and

- Alternate models are retained for an observation period while performance statistics are being collected (unlike SR, in which a new alternate model was generated each time period).

The agent applying SRF need only remember the information required to compare the current model and the alternate model; information necessary to evaluate other models will be required only when those models are enumerated.

If the agent is permitted to forget data, additional conditions are needed to determine the asymptotic behavior of SRF under the mixture distribution. Specifically, we need to know that the nature of the process is not changing with time: that the models which fit well now are likely to be the ones which fit well in the past. If this condition is not met, then SRF will be too likely (relative to SR) to accept models which fit poorly in the past but now fit well. This could cause SRF to be fooled into dropping the correct model in favor of one that temporarily predicts well, but will begin predicting poorly at some future time.[4]

I now specialize to a particular structure for the distribution of the $X_t$. Let $B_{1t}, ..., B_{qt}$ be a set of random variables observed by the agent before she observes $X_t$, and $A_{1t}, ..., A_{rt}$ a set of random variables observed by the agent after she observes $X_t$. The $B_{kt}$ and $A_{kt}$ are assumed to have only finitely many possible values. Suppose that under each $P_\omega$, the random vectors

$$S_t = (B_{1t}, ..., B_{qt}, X_t, A_{1t}, ..., A_{rt})$$

are exchangeable.[5] That is, the sequence $S_1, S_2, ...$ can be modeled as a set of independently and identically distributed random vectors with a common unknown probability distribution. The set $\Omega$ indexes a finite number of different model structures. For example, the different $\omega$ might specify different independence relationships among the variables comprising $S_t$. Each $P_\omega$ would include both a model structure and a procedure for estimating model probabilities given past data.

The SRF strategy compares models over a trial period of length k. The agent begins the $n^{th}$ trial period with a current model $\omega_{cn}$ and an alternate model $\omega_{an}$ (the initial values $\omega_{c1}$ and $\omega_{a1}$ may be chosen arbitrarily). She observes data over the $n^{th}$ trial period, which consist of the variables $S_{nk-k+1}, ..., S_{nk}$. The agent then computes the likelihood under each of the two models of the data observed during the $n^{th}$ trial period:

$$\lambda_{\omega n} = P_\omega(X_{nk-k+1} \mid \mathcal{H}_0, B_{nk-k+1}) \quad (11)$$
$$\cdot P_\omega(X_{nk-k+2} \mid \mathcal{H}_0, S_{nk-k+1}, B_{nk-k+2})$$
$$... P_\omega(X_{nk} \mid \mathcal{H}_0, S_{n-k+1}, ..., S_{nk-1}, B_{nk})$$

for $\omega = \omega_{cn}, \omega_{an}$. At the end of the $n^{th}$ trial period, the agent selects the current and alternate models for the $n+1^{st}$ trial period. The current model is replaced by the alternate model if the product of its $n^{th}$ period likelihood and its prior probability is higher than that of the current model:

$$\omega_{c,n+1} = \begin{cases} \omega_{cn} & \text{if } \alpha_{\omega_{cn}0}\lambda_{\omega_{cn}n} \geq \alpha_{\omega_{an}0}\lambda_{\omega_{an}n} \\ \\ \omega_{an} & \text{otherwise} . \end{cases} \quad (12)$$

The next alternate model $\omega_{a,n+1}$ is selected according to a probability distribution $\mu(\omega \mid Q_n)$, where $Q_n$ represents all the information available to the agent during the $n^{th}$ trial period:

$$Q_n = (S_{nk-k+1}, S_{nk-k+2}, ..., S_{nk}, \omega_{cn}, \omega_{an}). \quad (13)$$

While the agent compares models based only on data collected since the alternate model was enumerated,[6] the agent's forecast is based on all data observed since the current model was enumerated:

$$p_t = P_{\omega_{cm}}(X_t \mid \mathcal{H}_0, S_{mk-k+1}, ..., S_{t-1}, B_t) \quad (14)$$

where $\omega_{cn}$ was first enumerated as the alternate model on trial m of the model selection process, was adopted after the observation period, and has been the current model since.

If the $S_t$ are independent and identically distributed, then the random sequence $Q_n$ is a Markov chain with fixed transition probabilities.[7] If $\mu(\omega \mid Q_n)$ is positive for all $\omega$ and none of the $P_\omega$ assigns zero probabilities, then any state can be reached from any other state. Under these conditions, the chain possesses a stationary distribution (see, e.g., Billingsley, 1979). That is, if the process is permitted to evolve for a long period of time, the frequency with which it occupies state q approaches a limiting probability $\pi_q$. Thus, the SRF process does *not* converge to the correct probabilities. From any state (including states in which the current model has high

---

[4] If the environment is changing slowly, the correct model is not in the search space, $\Omega$ contains models which predict well in the short term, and the search process is likely to find a good short-term predictor when the current one begins to fail, then SRF may perform better than SR. Analyzing the non-stationary case would be an interesting extension to the present paper.

[5] The sequence $S_1, S_2, ...$ is said to be exchangeable if the $S_t$ have the same joint probability distribution when arbitrarily reordered.

[6] Restricting the agent to current period data greatly simplifies the mathematics. Again, no claim is made to optimality or fidelity to how real agents reason.

[7] The assumption I made was that the $S_t$ were exchangeable. This means they have the same distribution as an independent and identically distributed sequence with unknown probability distribution. Under exchangeability, $Q_n$ has the same distribution as a Markov chain with unknown transition probabilities. Thus, distributional results for Markov chains apply.

4164    Laskey

posterior probability relative to the other models) it is possible to reach any other state (including states in which the current model has very low posterior probability).

At equilibrium, transitions out of a state must balance transitions into the state. Let C denote the set of states for which the limiting posterior probability is greater than zero (these must all be asymptotically correct models), and D denote the set of states for which the limiting posterior probability is equal to zero (if there are any asymptotically incorrect models, D must be nonempty by Lemma 1). Let $p_{st}$ denote the probability of transition from state s to state t. Then

$$\sum_{\substack{s \in C \\ t \in D}} \pi_s p_{st} = \sum_{\substack{s \in C \\ t \in D}} \pi_t p_{ts} \qquad (15)$$

Increasing the length k of the trial period decreases the spread of the posterior probabilities about their limiting values. This decreases the probability of transition out of C and increases the probability of transition out of D. The effect of this is to increase the equilibrium probabilities of states in C. But increasing the sample size also increases the amount of time the system stays with each model, and therefore slows down the time it takes to move away from an initial poor-fitting model. This suggests the possibility of beginning SRF with a short trial period and gradually increasing the trial period as time goes on.[8]

In summary, the SRF model was analyzed on a restricted class of problems--those in which the probability of $E_t$ could be modeled as a time-invariant but unknown function of a finite set of auxiliary variables with finite sample spaces. It was shown that SRF with a fixed trial period does not converge to mixture model probabilities under the mixture model, but that increasing the length of the trial period decreases the probability of states in which the current model probabilities are far from the mixture probabilities.

## 4. INCORRECT MODELS

The foregoing discussion assumed that the mixture model was adequate. That is, the agent's larger world probabilities are the same as a model in which there is an unknown correct model in the agent's search space, and the agent believes that the mixture model probabilities will eventually match empirical frequencies arbitrarily closely. This section considers what can be said about the search processes defined above when the agent takes seriously the proposition that none of the models is asymptotically correct.

The behavior of a mixture distribution when none of the models is correct depends on the limiting behavior of the model likelihoods under the correct distribution. If $\Pi_t(X_t)$ is the actual probability of $X_t$, then it may be reasonable to assume that a version of the law of large numbers holds, and that under $\Pi$,

$$U_T = \frac{1}{T} \sum_{t=1}^{T} \log P_\omega(X_t \mid \mathcal{H}_t) \qquad (16)$$

approaches a limit as $T \to \infty$. This will occur, for example, in the exchangeable variables model assumed in Section 3.3 (assuming that none of the models assigns zero probabilities).

The average log-likelihood $U_T$ is a quantity often used to evaluate probability forecasts. Suppose a forecaster assesses probabilities over a set of events, and when the observation is made receives a score log(p), where p is the probability assigned to the event that actually occurred. The value $U_t$ is the agent's average score under this *logarithmic scoring rule*. The logarithmic scoring rule is an example of a *proper* scoring rule: that is, the expected score is maximized by assigning the correct probability distribution.

Note that if two models have equal prior probabilities, SR selects the model for which the past logarithmic score is highest. As T approaches infinity, the effect of the prior wears off; in the limit, SR settles on the model with the highest logarithmic score. A future paper will consider the performance of SR under different scoring rules, and considers a modified SR that converges to the model whose asymptotic score is highest under scoring rules other than logarithmic.

## A APPENDIX: PROOFS OF RESULTS

The proofs of Lemma 1 and Corollary 1 do not depend on finiteness of $\Omega$.

*Proof of Lemma 1*: Let

$$\lambda_{\omega t} = P_\omega(X_1 \mid \mathcal{H}_0) P_\omega(X_2 \mid \mathcal{H}_1) \cdots P_\omega(X_t \mid \mathcal{H}_{t-1})$$

be the likelihood of the observed data under the model indexed by $\omega$. Consider the sequence of random variables

$$R_t(\omega, \omega^*) = \frac{\lambda_{\omega t}}{\lambda_{\omega^* t}} \qquad (A-1)$$

defined by ratios of likelihoods of the data under $P_\omega$ and $P_{\omega^*}$, respectively. This likelihood ratio satisfies the equation $E[R_t(\omega,\omega^*) \mid \mathcal{H}_{t-1}] = R_{t-1}(\omega,\omega^*)$, where the expectation is taken under the distribution $P_{\omega^*}$. A sequence with this property is called a martingale. Because the martingale (5) is always nonnegative, the martingale convergence theorem applies (cf., Billingsley, 1979). Therefore, there exists a limiting random variable $R(\omega,\omega^*)$ such that $R_t(\omega,\omega^*) \to R(\omega,\omega^*)$ with probability 1 under $P_{\omega^*}$.

The random variable

$$R_t(\omega^*) = \sum_{\omega \in \Omega} \alpha_{\omega 0} R_t(\omega, \omega^*) \qquad (A-2)$$

---
[8] It should be possible to prove that a suitably chosen schedule for lengthening the trial period implies convergence in probability, but that is a matter for another paper.

is also a nonnegative martingale, and therefore also converges to a limiting variable $R(\omega^*)$. Note that $R_t(\omega^*) \geq \alpha_{\omega^*0} R_t(\omega^*,\omega^*) = \alpha_{\omega^*0}$, so that $R_t(\omega^*)$ is bounded away from zero. The posterior probability of model $\omega$ is given by

$$\alpha_{\omega t} = \frac{\alpha_{\omega 0} R_t(\omega,\omega^*)}{R_t(\omega^*)} \ . \qquad (A-3)$$

This ratio converges to a limit $\alpha_\omega$ equal to the quotient of the limits of numerator and denominator. From (A-1) it is clear that $R_t(\omega,\omega^*)$ converges to a nonzero limit, and therefore that $\alpha_\omega > 0$, only if (5) holds.

*Proof of Corollary 2*:

$$| P(X_t | \mathcal{H}_{t-1}) - P_{\omega^*}(X_t | \mathcal{H}_{t-1}) |$$

$$\leq \sum_{\omega \in \Omega} \alpha_{\omega t} | P_\omega(X_t | \mathcal{H}_{t-1}) - P_{\omega^*}(X_t | \mathcal{H}_{t-1}) |.$$

The quantities $| P_\omega(X_t | \mathcal{H}_{t-1}) - P_{\omega^*}(X_t | \mathcal{H}_{t-1}) |$ are bounded above by 1, and $\alpha_{\omega t}$ converges to zero unless (A-4) holds. But (A-4) implies that $| P_\omega(X_t | \mathcal{H}_{t-1}) - P_{\omega^*}(X_t | \mathcal{H}_{t-1}) | \to 0$. Therefore, each term in the sum converges to zero, which implies that the sum converges to zero.

*Proof of Theorem 3*: The proof of Lemma 1 established that $\alpha_{\omega t} \to \alpha_\omega$ with probability 1. Select the index $\omega_1$ so that $\alpha_{\omega_1} \geq \alpha_\omega$ for all $\omega \in \Omega$.

Note that if all $\alpha_\omega$ are equal to $\alpha_{\omega_1}$ then (5) holds for all $\omega$ (by Lemma 1) and the theorem holds. Therefore, we need only consider the case in which some $\alpha_\omega$ is strictly less than $\alpha_{\omega_1}$. Select $\omega_2$ as index of the next largest $\alpha_\omega$ not equal to the first: $\alpha_{\omega_2} = \max \{\alpha_\omega : \omega \in \Omega$ and $\alpha_\omega \neq \alpha_{\omega_1}\}$.

Let $q = \frac{1}{2}(\alpha_{\omega_1} - \alpha_{\omega_2})$. For each $\omega$ there is an index $T_\omega$ such that $| \alpha_{\omega t} - \alpha_\omega | < q$ if $t > T_\omega$. Let $T = \max\{T_\omega\}$. Then for $t > T$, $| \alpha_{\omega t} - \alpha_\omega | < q$ for all $\omega \in \Omega$. This implies that

$$\alpha_{\omega_1 t} - \alpha_{\omega t} > \alpha_{\omega_1} - \alpha_\omega - 2q > \alpha_{\omega_2} - \alpha_\omega \ .$$

This latter quantity is nonnegative unless $\alpha_\omega = \alpha_{\omega_1}$. Thus, once an $\omega$ for which $\alpha_\omega = \alpha_{\omega_1}$ is visited on a trial greater than T, it will remain the case that $\alpha_{\omega_{ct}} = \alpha_{\omega_1}$ on all future trials. That is, $\omega_{ct}$ will remain at $\omega$ for which the limiting posterior probability is strictly positive. These are values for which (5) holds. Hence, (8) holds. The foregoing argument is valid for each realization for which the $\alpha_{\omega t}$ all converge, and Lemma 1 established that such realizations occur with probability 1 under $P_{\omega^*}$.

*Proof of Corollary 4*: Suppose the agent believes that the $X_t$ actually arise from one of the distributions $P_{\omega^*}$. By Theorem 3, the agent believes that (8) holds, which implies that $| P_{\omega_{ct}}(X_t | \mathcal{H}_{t-1}) - P_{\omega^*}(X_t | \mathcal{H}_{t-1}) | \to 0$ with probability 1. By Theorem 2, $| P(X_t | \mathcal{H}_{t-1}) - P_{\omega^*}(X_t | \mathcal{H}_{t-1}) | \to 0$ with probability 1. Therefore, $| P_{\omega_{ct}}(X_t | \mathcal{H}_{t-1}) - P(X_t | \mathcal{H}_{t-1}) | \to 0$, with probability 1. But probability 1 convergence implies convergence in distribution, which is (9). But if (9) holds for any process distributed according to the mixture distribution P, it holds for all such distributions.

### Acknowledgments

This work was supported in part by a grant from the Virginia Center for Innovative Technology to the Center of Excellence in Command, Control, Communications and Intelligence at George Mason University. The author thanks Paul Lehner for many helpful discussions and an anonymous reviewer for insightful comments on an earlier draft of the paper.

### References

Andreassen, S., Woldbye, M., Falck, B. and Andersen, S.K., MUNIN: A causal probabilistic network for interpretation of electromyographic findings. In *Proceedings of the Tenth International Joint Conference on Artificial Intelligence*, 366-372, 1987.

Billingsley, P. *Probability and Measure*. New York: Wiley, 1979.

Goldman, R.P. and Charniak, E. Dynamic construction of belief networks. In *Proceedings of the Sixth Conference on Uncertainty in Artificial Intelligence*, 90-97, 1990.

Howsen, C. and Urbach, P., *Scientific Reasoning: The Bayesian Approach*. LaSalle, IL: Open Court, 1989.

Kuhn, T. *The Structure of Scientific Revolutions*. Chicago: University of Chicago Press, 1962.

Laskey, K. B. Conflict and Surprise: Heuristics for Model Revision. In D'Ambrosio, B.D., P. Smets and P.P Bonissone (eds.). *Uncertainty in Artificial Intelligence: Proceedings of the Seventh Conference (1991)*. San Mateo, CA: Morgan Kaufmann, 1991.

Pearl, J. *Probabilistic reasoning in intelligent systems: Networks of plausible inference*. San Mateo, CA: Morgan Kaufman Publishers, Inc., 1988.

Wellman, M.P. and Heckerman, D.E., The role of calculi in uncertain reasoning. *Proceedings of the Third Workshop on Uncertainty in Artificial Intelligence*, 321-331, Seattle, WA, July 1987.